\documentclass{article}
\usepackage{graphicx} 

\usepackage[
backend=biber,
style=alphabetic,
sorting=ynt
]{biblatex}

\usepackage{amsmath}
\usepackage{subcaption}

\title{Deep Semi-Supervised Anomaly Detection for Finding Fraud in the Futures Market}

\author{Timothy DeLise\thanks{Département de mathématiques et de statistique, Université de Montréal\newline Acknowledgment: This research was funded by the Graduate Excellence Scholarship of the Centre d’intelligence en surveillance des marchés financiers (CISMF) de l’UQAM in the summer of 2021. Additional thanks to the TMX Group, our private sponsor and collaborator in this research for providing data, insight and context.}}
\date{\today}
\date{July 2023}

\begin{document}

\maketitle

\begin{abstract}

Modern financial electronic exchanges are an exciting and fast-paced marketplace where billions of dollars change hands every day. They are also rife with manipulation and fraud. Detecting such activity is a major undertaking, which has historically been a job reserved exclusively for humans. Recently, more research and resources have been focused on automating these processes via machine learning and artificial intelligence. Fraud detection is overwhelmingly associated with the greater field of anomaly detection, which is usually performed via unsupervised learning techniques because of the lack of labeled data needed for supervised learning. However, a small quantity of labeled data does often exist. This research article aims to evaluate the efficacy of a  deep \textit{semi-supervised} anomaly detection technique, called Deep SAD, for detecting fraud in high-frequency financial data. We use exclusive proprietary limit order book data from the TMX exchange in Montréal, with a small set of true labeled instances of fraud, to evaluate Deep SAD against its unsupervised predecessor. We show that incorporating a small amount of labeled data into an unsupervised anomaly detection framework can greatly improve its accuracy.  
    
\end{abstract}

\section{Introduction}
\label{section:intro}

Fraud detection in high frequency data is one important aspect of forensic work performed by the oversight teams at derivatives exchanges worldwide. Each day millions of lines of orders and transactions are recorded in real time. Certain technically destructive behaviors have been identified, such as \textit{layering} or \textit{spoofing} \cite{Gideon:2020}, but other more subtle types exist such as insider trading. It is certainly a complicated task, often with legal implications associated. Given the importance and sensitivity of the job, the painstaking task of identifying instances of fraud has historically been left up to humans.

Amidst the recent clamour around topics like artificial intelligence and machine learning, in addition to the proliferation of high frequency market data, the Toronto Mercantile Exchange's (TMX) derivatives trading operation, located in Montreal, Quebec, has has taken the initiative, inspired by recent developments in ML and AI, to research automated ways that technology can help in the fight to uphold the integrity of financial markets in Canada and worldwide.

Financial institutions often have a lot of data, such as order book data, or credit card transaction data, but they do not necessarily know if each data point was an instance of a fraud (or anomaly) or not. The process of flagging such data and finding out is a very slow and burdensome. After years of investigating, a team may only have found a handful of confirmed instance of fraud, while the size of data is in millions or billions of transactions. To be fair, most transactions we assume are normal, with fraudulent or anomalous data representing only a small portion of all the data. Fraud detection in financial data has been dominated by the the field of unsupervised learning \cite{GoldsteinUchida:2016, app13105916}. Unsupervised learning techniques aim to find models which can learn and explain the latent structure of data\cite{GoodBengCour16, Makhzani:2018}. They do not require labeled targets as in supervised learning problems. This is one of their main advantages to the field of fraud detection. Without labeled data, it is also one of the only options.

One of the most popular unsupervised techniques for fraud detection in financial data using ML and AI are deep neural network autoencoders \cite{DBLP:journals/corr/abs-1908-11553}. An autoencoder model's job is simple. We provide our data as input, and the model simply needs to recreate that exact data as best it can. We impose limitations on the model, such as a low-dimensional bottleneck in the neural network, which forces a low-dimensional representation of the input data \cite{rumelhart:errorpropnonote}. For each input data point, the model produces an output, and training is performed via gradient descent-style algorithms to minimize the mean-squared distance between input and output. No model is perfect, and there are still non-zero errors associated with each data point after training. The magnitude of the errors can be a measure of how different, or anomalous, a data point is. A large error means that the autoencoder chose to sacrifice performance on that data in favor of other data. It means that this data is out-of-the-ordinary. Auto-encoder-based anomaly detection programs associate training error with \textit{anomaly score}. The higher the score, the greater chance that the data point is an anomaly. Fraud detection schemes then associate the idea of \textit{anomaly} with \textit{fraud}. It is a loose specification and teams should make sure to realize that a high anomaly score is not alone evidence of a crime.

The fact that we often do have a few examples of fraud (labeled data) has made the idea of a \textit{semi-supervised} approach to fraud detection very enticing. Indeed it has been the subject of recent research \cite{VILLAPEREZ2021106878}, and it makes sense. Any amount of labeled data, however small, is valuable.  The authors of \cite{Goernitz_2013} point out that the performance of unsupervised approaches often too low to be useful, but purely supervised technique fail to generalize well to unseen new kinds of anomalies. \textbf{They recommend that any semi-supervised anomaly detection system should be grounded in the unsupervised paradigm.}

The TMX derivatives exchange had already been working with an unsupervised model similar to the Deep Support Vector Data Description (Deep SVDD) model\cite{pmlr-v80-ruff18a}. The formulation of this model is a slight deviation from the traditional autoencoder. Instead of training the deep neural network to recreate the input data, the network is trained to minimize the volume of a hypersphere that encloses the network representation of the data. This effect is realized through a novel objective loss function for this purpose, in line \ref{eq:svdd} below. In this expression $x_i$ is the $i^{\text{th}}$ data point out of $n$ total data points, $\phi$ is the neural network model function, $\mathcal{W}$ are the weights of the neural network to be optimized, and $\textbf{c}$ is the center of the hypersphere.

\begin{equation}
\label{eq:svdd}
    \min_{\mathcal{W}} \frac{1}{n} \sum_{i=1}^n || \phi(x_i;\mathcal{W}) - \textbf{c}||^2
\end{equation}

The model has shown promise, but there remains a question around the small set of labeled anomalies (frauds) that we have. Can we further leverage this data to improve model performance? The purpose of this current paper is to investigate the efficacy of moving to a semi-supervised paradigm. To this end, we explore a further development, from the authors of the Deep SVDD model, called Deep SAD, which stands for \textit{Deep Semi-Supervised Anomaly Detection}\cite{DBLP:journals/corr/abs-1906-02694}. This is a method, grounded in the unsupervised learning paradigm, which takes advantage of an arbitrary number of labeled examples alongside the vast amount of unlabeled data. This is accomplished by introducing a new term into the objective function on line \ref{eq:svdd} above. The new objective function minimizes the distance of the neural network representation from the center of the hypershpere for the unlabeled data while maximizing the same distance for the small set of labeled true anomalies. Normal data should be toward the center, and anomalous data should be toward the outside. This new objective function is written here in equation \ref{eq:sad}.

\begin{equation}
    \label{eq:sad}
    \min_{\mathcal{W}} \frac{n}{n+m} \sum_{i=1}^n || \phi(x_i;\mathcal{W}) - \textbf{c}||^2 + \frac{\eta}{n+m} \sum_{j=1}^m \left( || \phi(\tilde{x}_j;\mathcal{W}) - \textbf{c}||^2 \right)^{\tilde{y}_j}
\end{equation}

In addition to the original $n$ unlabeled data points, $x_i$, we also have $m$ labeled data points $\tilde{x}_j$ with associated labels $\tilde{y}_j \in \{-1,1\}$. Where $\tilde{y}_j = 1$ denotes normal data points and $\tilde{y}_j = -1$ denotes anomalous samples. We see that when the labeled data is from normal data, the new term just replicates the job of the old term. But, for the anomalous data, the negative in the exponent induces a maximization of the distance for those data points. It should be noted here that the L2 penalty weight decay term has been omitted from the previous loss function for clarity. In practice there is an additional L2 penalty on the magnitude of the weights of the deep neural network in these objective functions.

Finally, the anomaly score, $\textbf{s}(x)$, of a data point $x$ is thus the distance of that data point's neural network representation from the center of the hypersphere.

\begin{equation}
    \label{eq:anomaly-score}
    \textbf{s}(x) = || \phi(x;\mathcal{W}) - \textbf{c} ||
\end{equation}

The empirical research to follow is aimed to confirm that Deep SAD indeed improves anomaly detection scores on a concrete internal data set provided by TMX. The following section outlines the data description and experimental methods. This is followed by the section with the results, and finally we conclude with a discussion of the findings.

\section{Methods}

\subsection{ Data Description }

Our data set consists of high-frequency limit order book data \cite{OHARA2015257} from one product of the futures market exchanges operated by TMX. The time duration of the data is is one month. This is to say that we have have every limit order book update to this particular exchange-traded financial asset over a one month time span. This totals 5,220,091 data points. Each data point contains 20 numerical fields representing the top 10 levels of the order book. Each level of the order book contains data for the price and size of the orders. This is one of the de facto standard data formats that are generated and distributed by electronic exchanges worldwide. For a detailed description of limit order books and their associated data, please refer to this resource \cite{RePEc:taf:quantf:v:13:y:2013:i:11:p:1709-1742}.

\begin{figure}
    \centering
    \includegraphics[width=\textwidth]{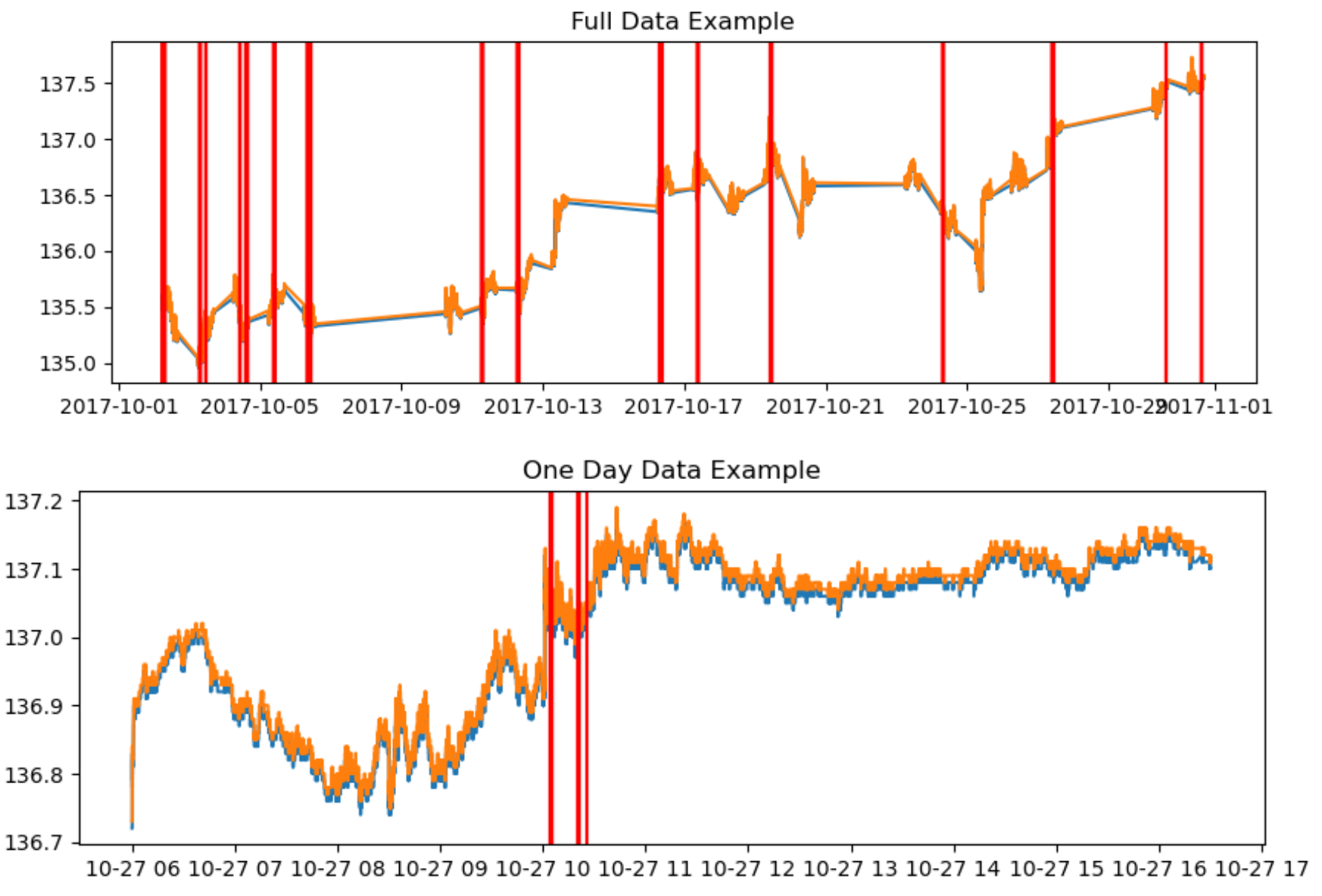}
    \caption{This figure helps describe the layout of the data. We plo the top two order book price levels over time. We see the dates vaary over the month of October 2017. The price varies between 135 to 137.5 dollars over this time. The red lines indicate the location of the true anomaly labels. The top pane iis the entire data set and the bottom pane zooms in on one day, October 27th. This helps understand how sparse the labeled data points are. }
    \label{fig:data-example}
\end{figure}

Out of the 5,220,091 available data points, 293 of them are labeled, and all the labeled data are true anomalies. In the context of the objective function above, which we will use later on, $n = 5,219,798$, $m = 293$, and $\tilde{y}_j = -1\  \forall j \in \{1, 2, ..., m\}$. The data set is visualized temporally in figure \ref{fig:data-example}.

\subsection{ Experimental Design }

For the experiment we utilize randomly initialized 3-layer feed-forward neural networks with 100 nodes per hidden layer. The input layer is size 20 and so is the output layer, which equals the shape of the input data (20 dimensions). ReLU activation functions are used on the output of each layer, except the final output layer, which is just a linear output layer. This kind of neural network is also called a ReLU Net, and is a standard example of a deep neural network model with decent capacity, and the quality of universal function approximation\cite{Hanin_2019}. Both SVDD and DeepSAD undergo an unsupervised pre-training phase, where they are both trained for 1,000 epochs as an autoencoder, learning to approximate the input. The center of the hypersphere is then initialized as the mean output of the neural net over all the training data. Finally, the main training routine is run for 10,000 epochs. This is the stage where the SVDD and the Deep SAD models differ. The objective function for SVDD is defined in equation \ref{eq:svdd} and for Deep SAD in equation \ref{eq:sad} of the introduction. Once training is complete, the anomaly scores (equation \ref{eq:anomaly-score}) of the test data are then evaluated and stored for downstream analysis.

The data set is trained and evaluated in a 3-fold cross validation routine. During each fold, the model is trained on 2/3 of the data and evaluated on the remaining 1/3 of the data. The data is not shuffled and split by date, so each fold of data is a contiguous chunk of time. There isn't any time frame overlapping from one fold to another. The entire experiment is is run twice, each time on 3 folds of data. In the following sections, we display results from each of the 6 trials separately, which we believe helps to emphasize the findings. We display results both in-sample (training set) and out-of-sample (test set). This enhances transparency into the generalization ability of the models.

In order to compare the two techniques against each other, we've developed two criteria which we call the \textbf{\textit{ratio test}} and the \textbf{\textit{rank test}}. The ratio test displays the mean anomaly score of the labeled data divided by the mean anomaly score of the unlabeled data. The rank test shows the average rank of the true anomalies relative to all the data when sorted by anomaly score. For the ratio test, a score above 1 shows the model is working at all, and a better model will produce higher scores. For the rank test, a better model will produce lower scores, since we have ordered the data so that a rank of 1 is the best, with the highest score. 

\section{Results}

\begin{figure}
    \centering

    \begin{subfigure}[b]{0.45\textwidth}
        \centering
        \includegraphics[width=\textwidth]{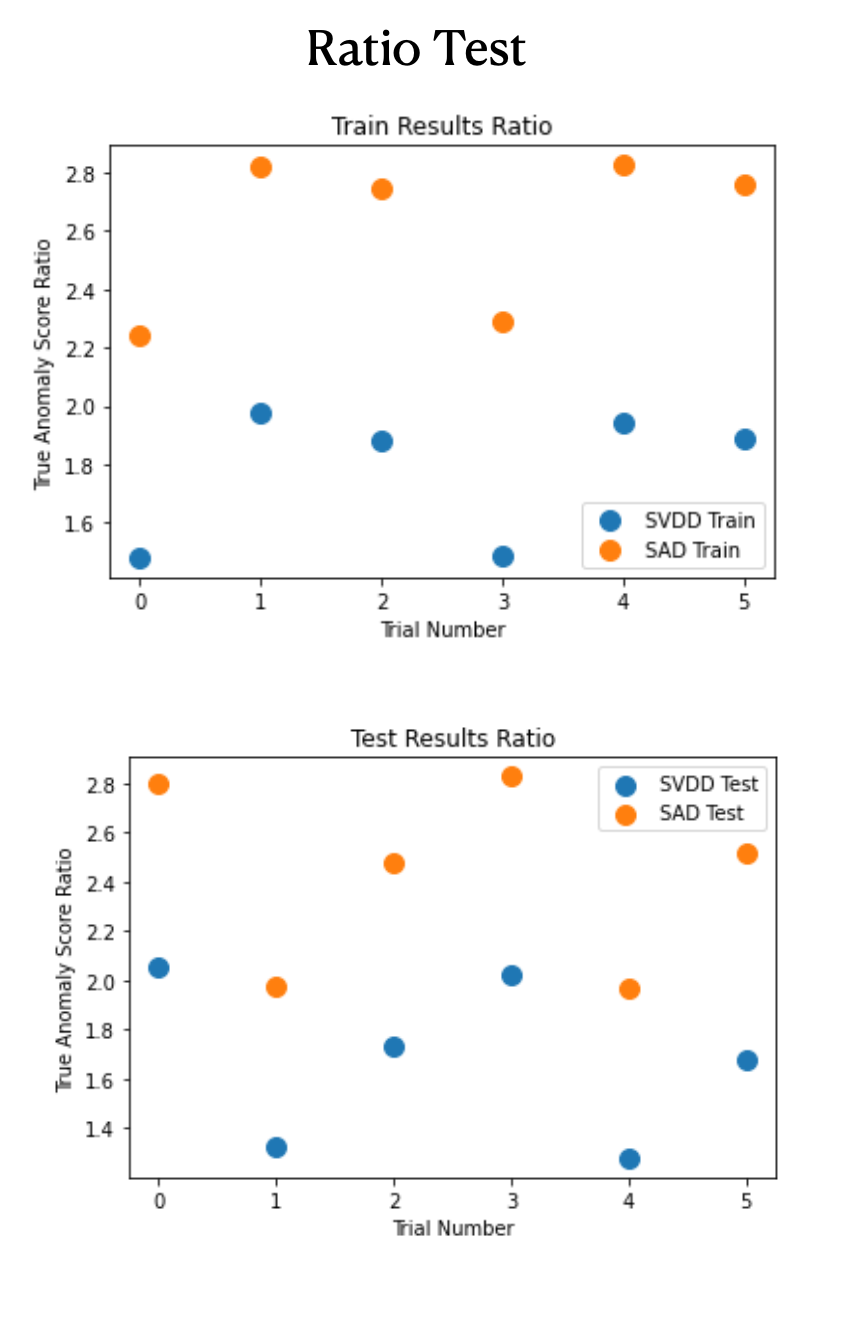}
        \caption{Results from the \textbf{Ratio Test}. In every trial, the ratio test score for Deep SAD is higher (better) than SVDD, both in-sample and out-of sample. }
        \label{fig:ratio-scores}
    \end{subfigure}
    \hfill
    \begin{subfigure}[b]{0.45\textwidth}
        \centering
        \includegraphics[width=\textwidth]{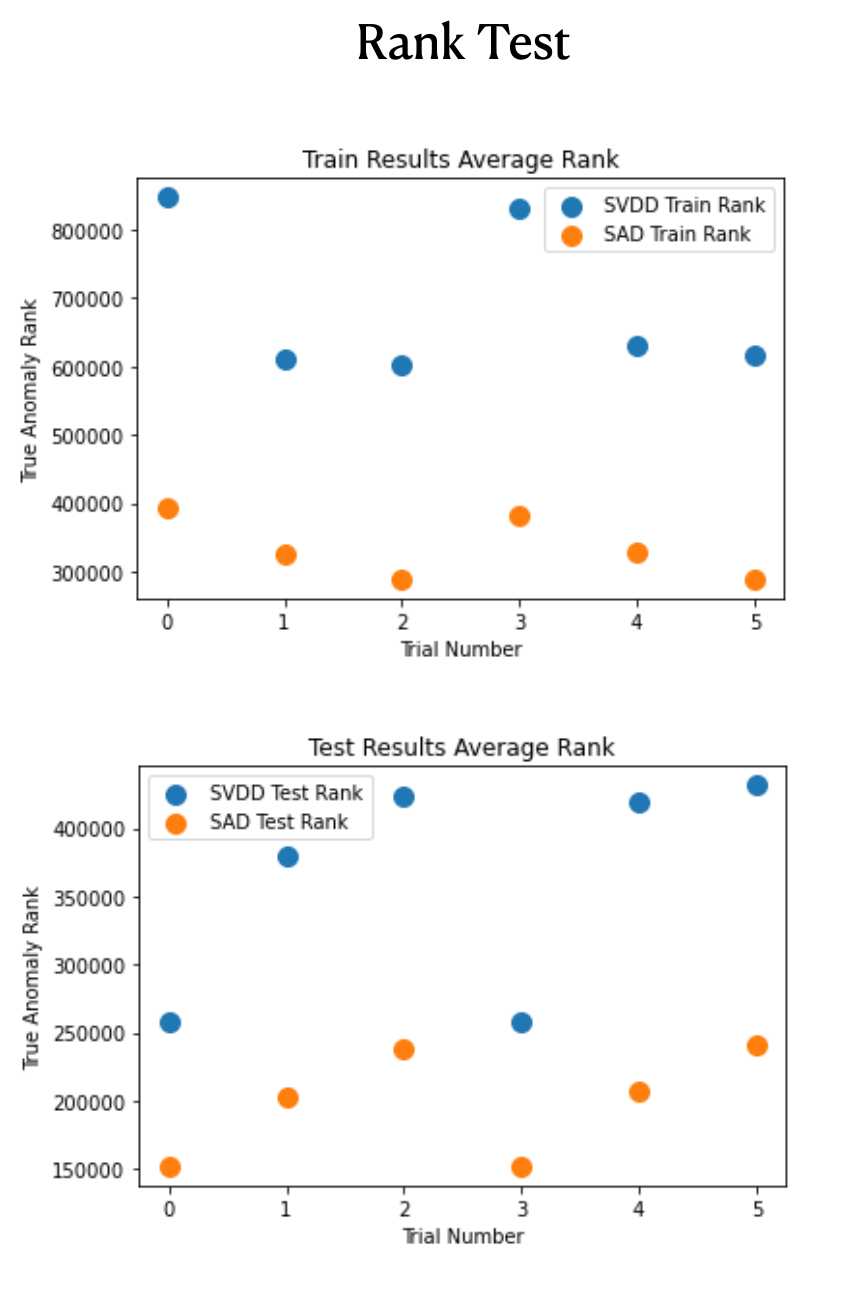}
        \caption{Results from the \textbf{Rank Test}. In every trial, the rank test score for Deep SAD is lower (better) than SVDD, both in-sample and out-of sample.}
        \label{fig:rank-scores}
    \end{subfigure}

    \caption{In each plot, the trial number is along the x-axis and the test score is along the y-axis. On the left side we have results from the ratio test and on the right side, results from the rank test. The top show in-sample results (train) and the bottom displays out-of-sample (test) results. Data points for the SVDD model are in blue and Deep SAD is in blue.}
    \label{fig:experiment-results}
\end{figure}

The results are quite clear and point to a convincing benefit of the Deep SAD model over the SVDD model. The results of the experiment are graphically shown in figure \ref{fig:experiment-results}.

The results of the ratio test show that, both in-sample and out-of-sample, the DeepSAD model attains a higher score. The scores range from 1.5 to 3. This means that the SVDD model, which scores at the lower end of the range, still attributes higher anomaly scores to the true anomalies as opposed to the unlabeled data. This is good. However, the Deep SAD model increases this ratio, so that the true anomalies score even higher compared to normal data. This shows that, by incorporating a small amount of labeled data into our workflow, we've greatly increased the performance of our anomaly detection system.

The rank test paints the same picture. We see that the Deep SAD model leads to better identify instances of true anomalies in the data, both in-sample and out-of-sample. We should clarify here that the lower the rank, the higher the score. The data point with rank zero would have the largest distance from the center of the hypersphere.

\section{Discussion}

\begin{figure}
    \centering
    \includegraphics[width=0.8\textwidth]{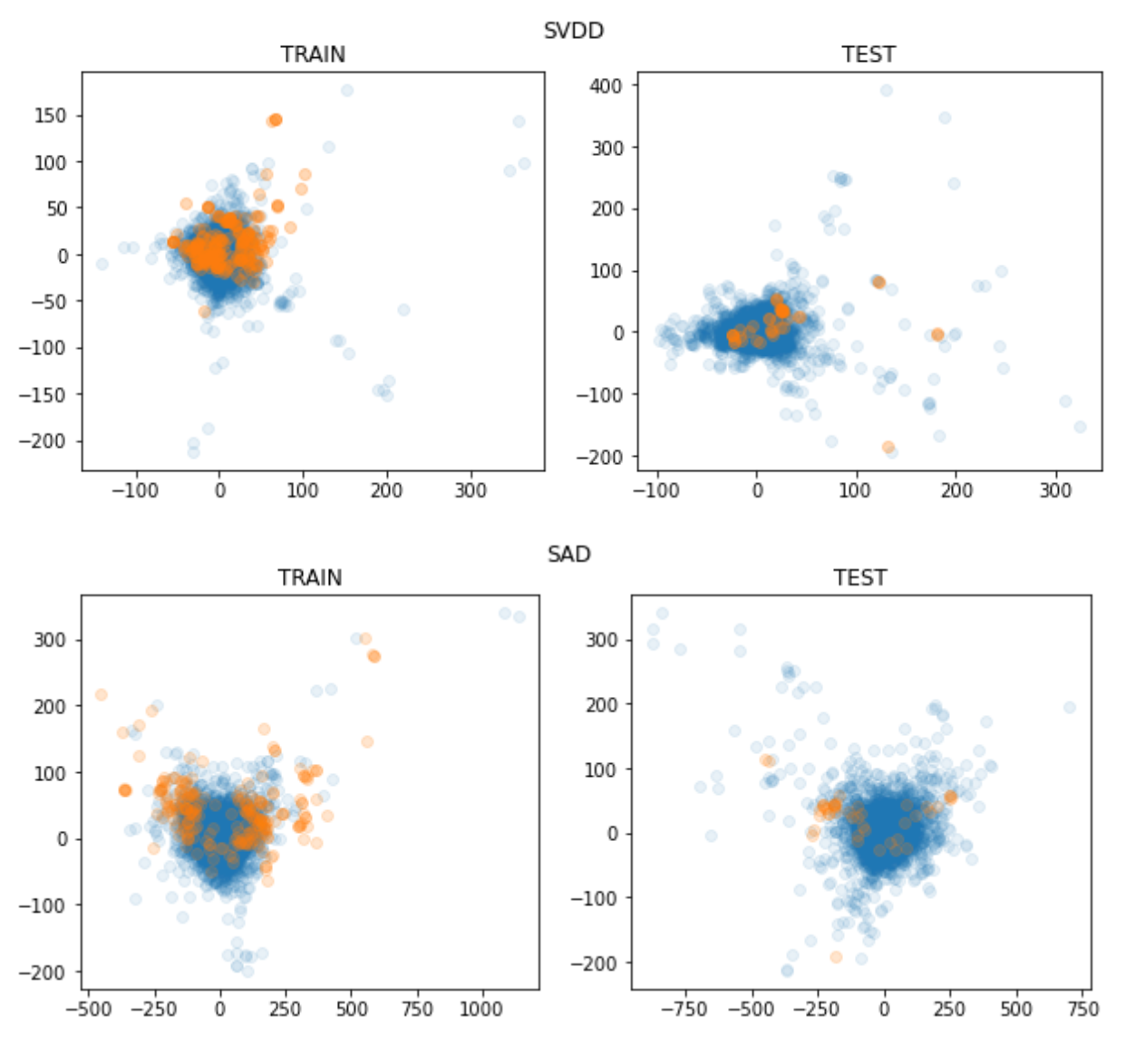}
    \caption{Visualizing the Distance to the Center of the Hypersphere. In this figure we plot the neural network representations of the data points after performing a principle component analysis to reduce the dimension of the data from 20 dimensions to 2. The blue dots are from unlabeled data and the orange dots are from the true labeled anomalies. There are far fewer labeled data points, so they are plotted on top. The results from the SVDD model, train and test set, are presented on the top row, while those for the Deep SAD model are presented on the second row. }
    \label{fig:viz-hypersphere}
\end{figure}

We've taken quite a journey during this research, and come a long way. We started by facing a seemingly insurmountable task of identifying fraud in the high-frequency financial markets, akin to finding a needle in a haystack. We've built on the ponderous ideas of the fore bearers of applied ML and AI. We've seen how unsupervised learning can form a basis to the fraud detection framework. Finally we've discovered here that we can effectively leverage a small amount of labeled anomaly (fraud) data to enhance our unsupervised pipeline.

At the crux of this new Deep SAD approach is idea that is baked into the objective function of the model; that the distance from the center of a hypersphere in high-dimensional space is an anomaly score. In order to get more intuition behind this, we've created a visualization which allows us to see the data points projected onto the space of the hypersphere, shown in figure \ref{fig:viz-hypersphere}. In this figure, each data point has been run through the neural network, and output data is compressed from 20 dimensional to 2 using principle component analysis. The blue data points are the unlabeled data, and the orange are the labeled true anomalies. For the SVDD model in the training set, we see the relative spread of the unlabeled data and the labeled data is about the same, while in the Deep SAD training set, the orange dots are pushed farther away from the center of the cluster. This visually confirms that, indeed, the Deep SAD training technique makes the true anomalies appear farther from the center of the hypersphere as opposed to the SVDD model. In the test sets, while there are less true anomalies, so the data is scarce, we see a similar pattern. For the SVDD model the true anomalies appear right near the center of the main cluster, while for the Deep SAD model, the true anomalies show up more spread out, farther away from the center.

This qualitative interpretation, along with the results from the previous section, suggest that this model generalizes surprisingly well. Affects in the training data transfer to the test data with minimal loss. 

Since this small amount of labeled data exists, it seems like quite a sacrifice to not incorporate it into an fraud detection pipeline for financial markets. We've shown here that semi-supervised frameworks like Deep SAD, grounded in the unsupervised paradigm and implementing deep learning, can benefit from both unlabeled data and a small amount labeled data. Surely there is value to creating and saving increasing amount of labeled true anomaly data. This research is hopeful to further guide students and practitioners to advance in this field of semi-supervised anomaly detection.

\medskip

\printbibliography

\end{document}